# RaspberryPI for mosquito neutralization by power laser

Rakhmatulin Ildar, PhD

South Ural State University, Department of Power Plants Networks and Systems

ildar.o2010@yandex.ru

**Abstract**

More than 700 thousand human deaths from mosquito bites are observed annually in the world. It is more than 2 times the number of annual murders in the world. In this regard, the invention of new more effective methods of protection against mosquitoes is necessary. In this article for the first time, comprehensive studies of mosquito neutralization using machine vision and a 1 W power laser are considered. Developed laser installation with Raspberry Pi that changing the direction of the laser with a galvanometer. A program for mosquito tracking in real-time has been developed. Given the calculations and results of experimental research. The possibility of using deep neural networks, Haar cascades, machine learning for mosquito recognition was considered. We considered in detail the classification problems of mosquitoes in images. A recommendation is given for the implementation of this device based on a microcontroller for subsequent use as part of an unmanned aerial vehicle. Any harmful insects in the fields can be used as objects for control.

## 1. Introduction

In the next papers used the developed device for remote mosquito control [15], for pest control [16] and for weed control [17]

The relevance of this work is the need to identify the presence of a mosquito in place and determine its position for subsequent destruction because mosquitoes carry a great danger because mosquitoes are carriers of diseases such as malaria, from which more than 200 million people in the world directly infect from mosquitoes. The danger and methods necessary to combat mosquitoes are described in the following articles [1,2,3,4].

In this paper, we propose to use machine vision, neural networks, and a 1-watt laser to neutralize mosquitoes. It is a new direction for control mosquitoes. For the first time, the idea of using a laser to protect against insects was expressed in the early 1980s by American astrophysicist Lowell Wood. But at the same time, due to the complexity of the technical implementation of this project, this idea has not yet found practical application. In the last decade, the rapid development of artificial intelligence technology, machine learning, and computer vision has provided an opportunity for the implementation of many ideas conceived decades earlier. Including the use of modern technology to combat mosquitoes.

The use of traditional methods such as nets, lotions, creams of various kinds, traps is especially popular as individual defense. Especially interesting works are [5,6,7,8]. A common disadvantage of all these methods that they are not autonomy.

As methods against mosquitoes, there are hundreds of different creams, devices, and other devices. But because of the analysis carried out, given that the problem with mosquitoes from year to year it becomes more and more relevant to conclude that no effective solution has yet been found to eliminate this problem. As a result, automating the process of neutralizing mosquitoes gives a favorable forecast for the expediency of using the method proposed in this paper.

In the past decade, the following methods have been proposed for mosquito control. In the article [9], the implementation of an acoustic insect flight detector, which was constructed from a noise reduction system, is presented, the response error was 6.5%. In our study, this detector can be used as an auxiliary method for determining the approximate area for the camera for work with mosquitoes. In the research [10], the classification of certain species of flying insects according to the frequency of beating of the wings is presented.

The following study is the most approximate in the topic of our work, where the results of mosquito detection using the camera are presented [11]. In this case, only one method was considered for detecting absolute difference estimation - and the results on determining the position of z are not presented.

It is worth adding that the Bill & Melinda Gates Foundation sponsored research on the implementation of laser mosquito neutralization technology as a promising method for combating malaria. As a result, company Photonic Sentry was created, information about this from the official site of the company was used. For several years of company operation, a demonstration video was presented on the neutralization of mosquitoes without a description of the installation.

The main advantage when using a laser to neutralize mosquitoes over known methods to combat mosquitoes are the following:

- ability to use in open areas, even in the presence of wind;
- the ability to create a given safe territory, which is protected from mosquitoes, around houses, or when traveling to nature around the camp;
- potentially - a large area of coverage;
- in the perspective, to protect a group of moving people who have left a safe territory, by mounting on an unmanned aerial vehicle or on the human body.

## 2. Materials and method

The equipment that was used in the research:
- Raspberry Pi 3 Model B +, Broadcom BCM2837B0 with a 64-bit quad-core processor (ARM Cortex-A53) with a frequency of 1.4 GHz;
- The program is written in python3.6, the library of machine vision OpenCV 3.4.1;
- Pi camera, Sony IMX219 Exmor;

- Galvanometer, speed - 20 kpps;
- Power laser -1 W, wavelength - 450 nm;
- From box with mosquitoes to laser system distance - 300 mm.

The size of a mosquito can vary from 1 mm to 5 mm, this is the main criterion for the method of detection and retrieval of mosquito coordinates. When monitoring the position of the mosquito with ultrasound, it is necessary to use several sensors in different places and processing their information to calculate the location, which theoretically is suitable only for detecting one mosquito, but if there are several mosquitoes, the device will not work correctly. The temperature of the body of a mosquito due to its cold blood is like the temperature of the environment. With a very high resolution of the thermal imager, the mosquito temperature will differ insignificantly against the background - on the order of 0.1 C. The use of sonar has several difficulties when working in open areas where it is necessary to use sonar with a narrow beam and a narrow radiation pattern.

Following the analysis, in the present study, it was decided to use a camera to detect and determine the coordinates of the mosquito. After the camera detects a mosquito, to increase the likelihood of a laser being affected it is necessary to predict the position of the mosquito in advance (by more than 0.2 s). Therefore, is necessary to research the factors affecting the behavior of the mosquito during flight.

This paper describes the effect of mosquitoes on odor [12]. In the paper [13], mosquito flight formulas are presented, which are calculated considering the presence of CO2. The speed of a mosquito was determined by a deterministic formula:

$$s = S_{max} - (S_{max} - S_{min}) * F(b, b_0) \qquad (1)$$

where $S_{max}$ $S_{min}$ – the maximum and minimum speed of a mosquito, F is a ramp function that takes values between 0 and 1.

In operation, the flight direction and speed of each mosquito are updated every ΔT time units. The updated position of mosquitoes is calculated by the formula:

$$\begin{pmatrix} x_{n+1} \\ y_{n+1} \end{pmatrix} = \begin{pmatrix} x_n \\ y_n \end{pmatrix} + s(\triangle T) * \vec{d} + \vec{V} * \triangle T \qquad (2)$$

where ($x_n$, $y_n$) - the position of the mosquito at time step n. Number d is a direction vector that varies between tracking and tracing. $\vec{V}$ – wind speed.

Considering the detection of a mosquito with the help of a camera, tracking it and using mathematical formulas to predict its further flight, the probability of positive control is increased.

To use a pre-trained deep learning network on the Raspberry Pi for the process of monitoring mosquitoes, initially, deep learning methods were considered to implement. But due to the limited RAM on the Raspberry Pi 3 (1 GB) and even the maximum configuration of the Raspberry Pi 4 (4 GB) and due to the low processor speed of 1.5 Ghz it is almost impossible to use the deep neural networks (ResNet> 100 MB, VGGNet> 550 MB , AlexNet> 200 MB, GoogLeNet> 30 MB). In an attempt to use SqueezeNet, using a new use of convolutions of 1 × 1 and 3 × 3, we managed to obtain a model with a weight of 5 MB, but even in this case, the result of image processing for the presence of the desired object was approximately 1 second.

Real-time detection with R-CNN, Fast R-CNN, Faster R-CNN, Yolo, RetinaNet has the same problem with speed of recognition. The solution may be to use - NVIDIA Jetson TX1 and TX2 - the special platform for computing neural networks. The main disadvantage, which is the high cost. Therefore, we focused on methods that can be implemented on the Raspberry PI, which allows you to create an economical and compact device.One of the most frequent ways to detect moving objects is the difference of frames, background subtraction, and analysis of the field of optical flow. Details about the advantages and disadvantages of these methods are written in paper [14].

Using the Hara cascade, you can train a model to track an object, the accuracy is determined largely by the number of photographs used to train the model. In the case found, 350 photos with mosquitoes were used as positive examples and 500 photos as negative examples. Detect mosquito by color using the cv2.cvtColor function. This function takes the original image and transforms the color space. In our case, HSV, RGB color spaces were considered. This method is more convenient for laboratory tests, in which you can create bright color contrast between the mosquito and the background, to study the dynamics of movement of mosquitoes and test the function of predicting the flight of a mosquito. In natural habitat due to the versatility of color, the effectiveness of this method is close to 0. The big focus of the camera will not improve the situation, because the mosquito itself does not have a single-color gamut, it is mean that for search in the program, it need expand the color ranges. In the end, it increases the noise.

Using optical flow like frame difference at high-resolution cameras can track the flight of a mosquito. In this case, since the image of visible movement represents the shift of each point between two images, all the color noise and noise present in the Hara cascade, in this case, does not matter. The disadvantage of this method is the impossibility of identifying the mosquito since the criteria, in this case, are only the size of the mosquito and the displacement of movement and analysis of the static image. In this case, one camera should cover a radius of 360 degrees, because of which the constant dynamics of the camera makes it impossible to use these methods.

The definition of an online location of an object in successive frames on a video is called tracking. In the OpenCV library through tracking, the following functions are considered: TrackerCSRT, TrackerKCF, TrackerBoosting, TrackerMIL, TrackerTLD, TrackerMedianFlow, TrackerMOSSE. Description of which are available on the website of the library developer. We have checked all the parameters. In the best way using - tracker = cv2.TrackerCSRT_create. The advantage of tracking is that it runs faster than detection. Since when an object that was detected in the previous frame is tracked, the program knows the initial data — the appearance of the object, the location, speed and direction of its movement. Fig. 1 shows the tracking of a mosquito with various methods, for mosquitoes size 2-4 mm.

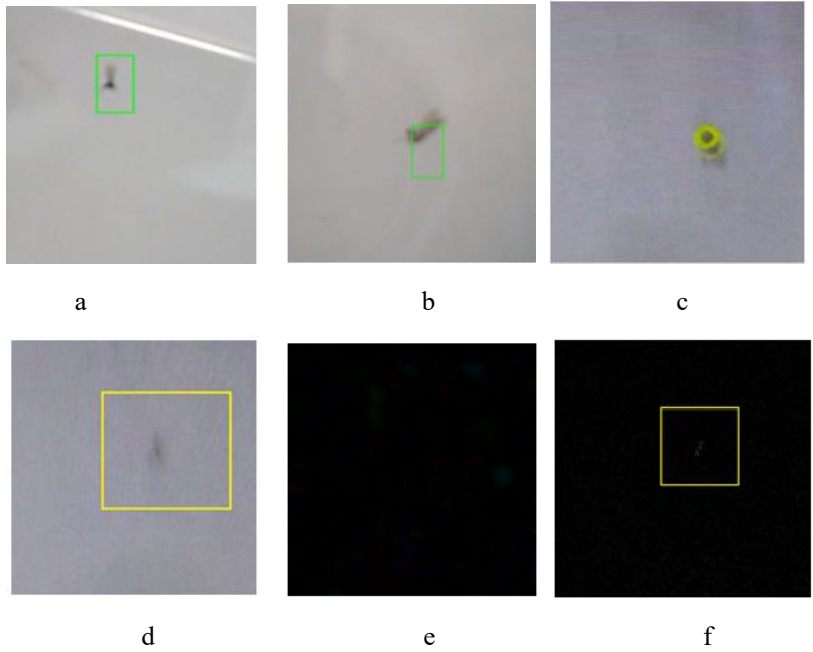

Fig.1. Using OpenCV for mosquitoes tracking: a – tracking mosquito with function cv2.TrackerCSRT_create (), b – Hara cascade, c – color tracking, d – color tracking without success, e – optical flow, f – frame difference

Fig.1.d shows a problematic situation for all the methods of tracking when working with mosquitoes, since with a sharp change of flight path the speed of the camera is not enough to fix the shape of a mosquito. Therefore, this moment all methods lose the mosquito from the field of view. Fig.1.e for the optical method with mosquitoes showed and many noises.

Using the image preprocessing function can significantly increase the readability of images for machine vision.

In Fig. 2 results in photo processing by Thresholding function-OpencV with different parameters.

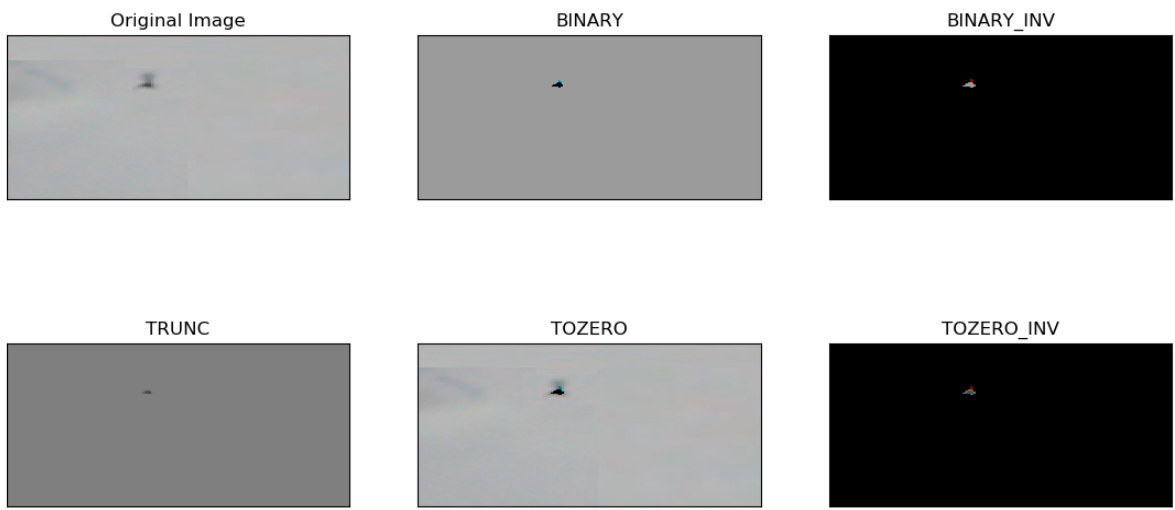

Fig. 2 results in photo processing by Thresholding function-OpencV

Image data can be used in classification problems, the author using convolutional neural networks obtained a classification accuracy of the order of-95%. For classification, a convolutional neural network (CNN) was created. It consists of 8 layers of layers Convolution_2D and MaxPooling2D after the second and fourth convolution. On all layers except the output fully connected layer, the ReLU activation function is used, the last level uses softmax. To streamline our model, a Dropout layer was used after each subsample layer and the first fully linked layer.

The network was trained on a stationary computer with an AMD Ryzen 5 3600 processor and a 6 GB GIGABYTE GTX1060 graphics card. The result obtained allows us to conclude that, in terms of classification, finding a mosquito in the images and calculating its exact coordinates is not difficult, but it takes time - more than 1 second. Classification can help in the case of the destruction of mosquitoes, which are not in motion since there is enough time in this case. But in this study, the task of neutralizing mosquitoes in dynamics is set.

To determine the distance to the object was used the stereo vision OpenCV function. At the same time, the depth value is inversely proportional to the displacement pixel, and the relationship between disparity and display depth of the mosquito, fig.3.

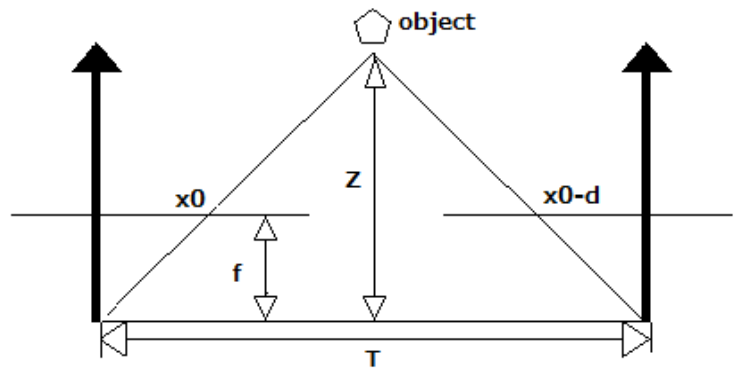

Fig.3. The layout of the cameras: d - a value called the disparity, Z - the depth or distance to the object, T - the distance between cameras, f - the coefficients of the focal length between the cameras.
The relationship between the camera and the object is described by the following formula:
$$(T-d)/(Z-f)=T/Z \qquad (3)$$

## 3. Experimental research

For experimental research, the installation was developed, Fig.4.

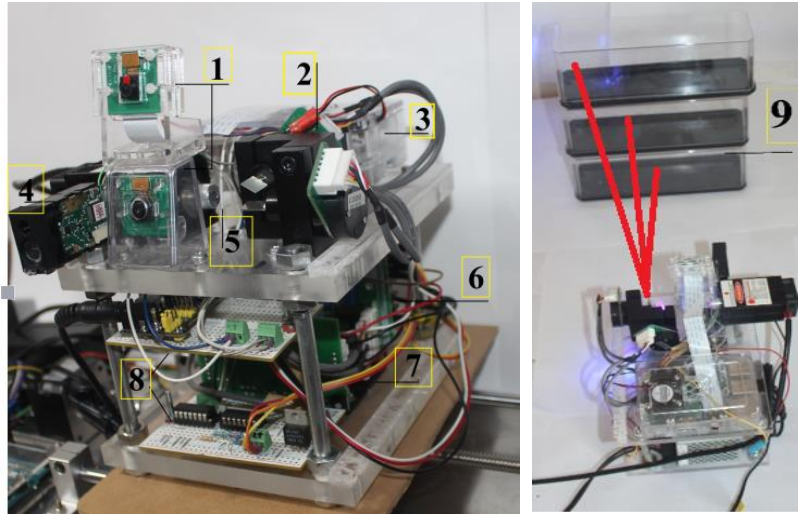

Fig.4. Laser installation: 1 - cameras, 2 - galvanometer, 3 - Raspberry PI3, 4 - laser rangefinder, 5 - laser, 6 - power supply, 7 - motor drivers, 8 - electronic signal processing board, 9 – box with mosquitoes

In the box 9 are mosquitoes. The laser and the camera are at a distance - 300 mm. The principle of operation of the installation is explained by the scheme in fig.5.

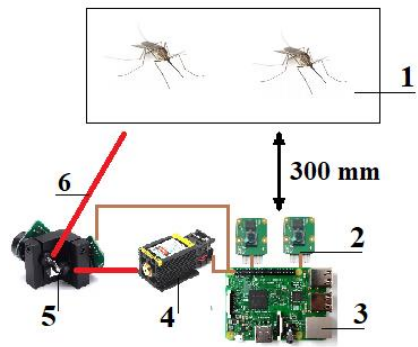

Fig.5. The scheme of the experiment: 1 - box with mosquitoes, 2 - pi cameras, 3 - Raspberry PI3, 4 - laser, 5 - galvanometer, 6 - laser beam

Raspberry PI 3 B + single-board computer, processes the digital signal from video and determines the position to the object and transmits a digital signal to the analog board - 3, where a digital-to-analog converter converts the signal to a range of 0-5V. Next, using a board with an operational amplifier, we get bipolar voltage - plus and minus 5 V, which feeds boards with a motor driver for a galvanometer - 4, from where the signal goes to galvanometers -7. The galvanometer with the help of mirrors changes the direction of the laser - 6. The system is powered by a power supply unit - 5. Camera 2 to determine the distance to the object.

Boxing with mosquitoes stands at a distance of 300 mm from the laser system. The camera finds the mosquito and transmits the data to the galvanometer, which puts the mirrors in the correct position, and then the laser is turned on.

The algorithm working installation is shown in fig.6.

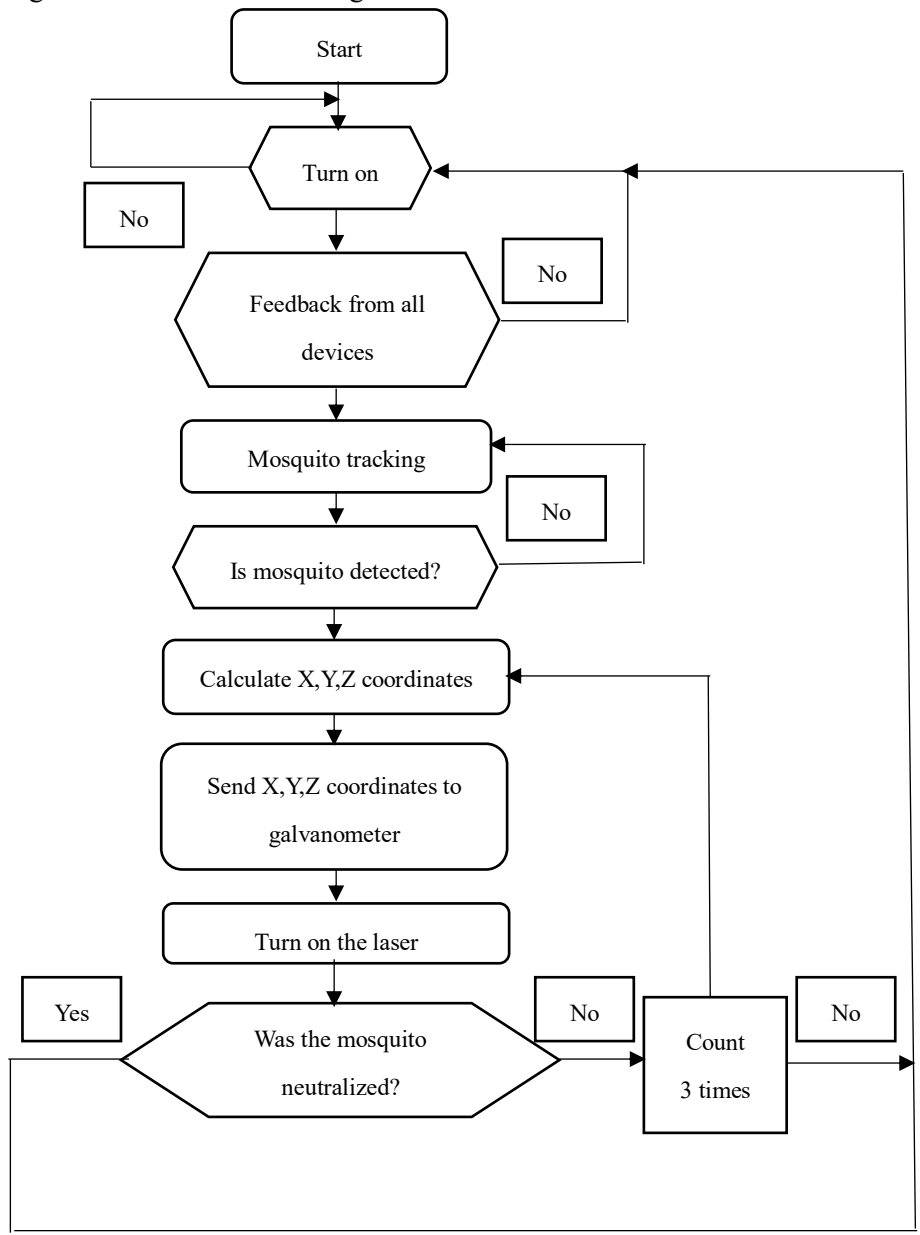

Fig.6. The algorithm working installation

The readings are averaged; the average is taken from 180 readings with different samples from the array
The results of experimental research using various OpenCV functions are shown in Table 1.

Table1. The results of experimental research

| Search method | The coordinates of the mosquito, X,Y, mm | Average time detect, sec | Tracking successful, % | Neutralization with a laser, %. With prediction formula. | Neutralization with a laser, %. Without prediction formula. | Survival after hitting a laser pulse – 0,5 s, % |
|---|---|---|---|---|---|---|
| By color | From 2,15 to 15, 20 | 0,3 | 65 | 13 | 10 | 55 |
| By tracking | From 0,18 to 10,12 | 0,15 | 76 | 15 | 13 | 56 |
| Cascades of haar | From 13,0 to 10,20 | 1 | 70 | 8 | 8 | 50 |
| Frame difference | From 0,18 to 20,20 | 0,1 | 62 | 3 | 5 | 75 |
| Optical flow | From 0,0 to 20,20 | ~ | ~ | 0 | 0 | ~ |

The developed prototype in this paper has a limited range of actions. But at the same time, the results of this work proved the possibility of using lasers to destroy mosquitoes. In the future, to increase the damaging ability, it is necessary to increase the accuracy of the laser operation.

To implement tracking in the Python language, various algorithms were written, both with tracking at each moment of only one mosquito and using the multithreading function and transmitting data on the position of mosquitoes using an array to the galvanometer. The success of the experiment can be enhanced by using a more powerful laser, which will make it possible to neutralize more than 2 mosquitoes in one second. Since considering the speed of the galvanometer, which allows you to change the thousandths of a position in seconds, we are limited in speed to neutralizing mosquitoes only by the power of the laser and the power of the central processor. The number of tracking mosquitoes at the same time depends on the processing power of the processor.

## 4. Discussion and conclusions

One of the factors limiting the use of laser technology to neutralize mosquitoes is the small allowable lesion area. The possibility of using a telephoto lens to monitor the position of mosquitoes was analyzed. In fig.7 prototype for remote monitoring of mosquito's present.

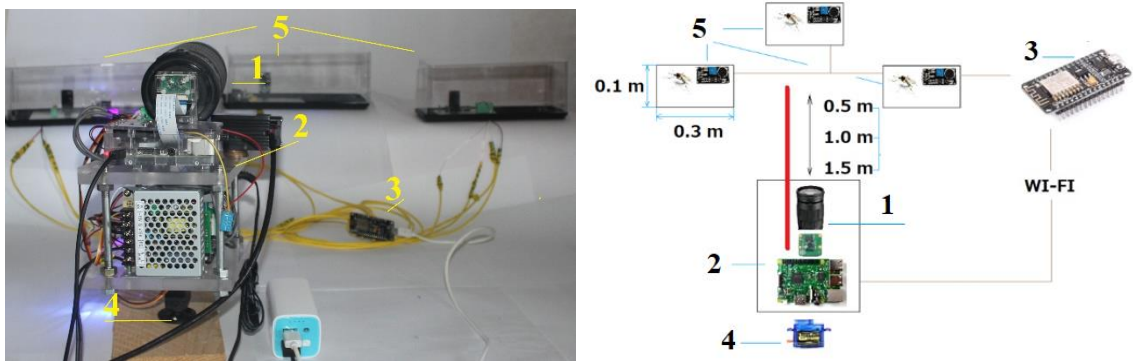

Fig.7. Prototype for remote monitoring of mosquitoes: 1 - telephoto lens, 2 - laser setup, 3 - microcontroller esp8266, 4 - servo motor to change the position along the axis-z 5 - three sections with mosquitoes and audio sensors.

The use of a telephoto lens and a servomotor for moving along the axis z as part of a laser installation can significantly increase the area of mosquito control. For regulating different distances to mosquito, we used - Esp8266, whereby WI-FI with standard IEEE 802.11 connection between boxes with mosquitoes and audio sensors and with Raspberry pi3 was realized. The distance of mosquito's flight monitoring depends on the resolution of the telephoto lens.

A promising development direction for the developed installation is the use of STMicroelectronics microcontrollers from the STM32 family. Given that in X-CUBE-AI - AI has an expansion pack for STM32CubeMX. This expansion can operate with different deep learning frameworks such as Caffe, Keras, TensorFlow, Caffe, ConvNetJs, etc. Due to this, the neural network can be trained on a stationary computer with the possibility of calculations on the GPU. After the integration to use an optimized library to a 32-bit microcontroller STM32.

At the same time, the use of a microcontroller instead of a single-board computer will significantly reduce power consumption by several watts. It is worth considering various programming languages, for example, a program on C will be more efficient, than python language.

As drivers for motors, it is recommended to use drivers based on SMD components. Like a rechargeable battery, for example, Thunder Power Adrenaline Weight not more than 179.69g and capacity 1100mAh This will significantly reduce the dimensions of the device and reduce its weight to 450 grams.

What ultimately increases the efficiency of the device, because it will be possible to use the device using an autonomous aircraft. Nowadays many companies specialize in the production of an apart capable of bearing the burden and fly significant distance without recharging.

The results of these works will be fully disclosed in subsequent research.

In the future necessary to use a laser with focusing. In this research, due to the insignificant distance between the laser and the mosquito box, the pre-configured focus of the laser did not play a big role, since the box width was 70 mm. The lesion area on the near side of the box exceeded the mosquito damage area by 2% by the distant side in the view. But it can be of the general error, in this research we do not focus on this.

It is advisable to consider replacing the audio sensors that were used to demonstrate the ability to neutralize mosquitoes over considerable distances. As a device for determining the distance can be used a laser rangefinder with a high data rate, where for an interface to use SPI. Since the mirrors can change position within milliseconds, monitoring with a laser rangefinder from technical implementation does not carry any difficulties.

Due to the mosquito's flight speed, 1 meter per second, with increasing distance from 15 meters the rotation speed along the z-axis with a servomotor greatly affects the result. Given this, recommend using the algorithm is implemented in such a way that after identification, mosquitoes are assigned numbers, if the mosquito has left the camera's observation area, its coordinate is calculated using the mathematical model and the laser focus is transferred to the calculated coordinates

In this case, this will give a significant increase in time. Otherwise since due to the mosquito leaving the observation zone, to determine the exact position, it is necessary to process the entire picture from the camera once again after turning along the z-axis, which is very expensive from time.

The use of neural networks to predict the position of a mosquito will lead to the installation of an additional microcontroller, which will lead to a 2-fold increase in the mass of equipment and the consumption of electrical energy.

In this research the idea of using a galvanometer with a powerful laser to neutralize mosquitoes was first realized. The system can neutralize 2 mosquitos per sec and this result can be easily improved. Experimentally proved the feasibility of using a laser for the destruction of mosquitoes. Developed a laser unit that changes the direction of the laser with a galvanometer. A program has been developed for the monitoring of mosquitoes in real-time. Calculations and results of experimental studies are given. Experimental research various techniques for mosquitoes and selected the best option for neutralizing mosquitoes with using in Raspberry PI - library OpenCV with function cv2.TrackerCSRT_createm.

**Funding Statement**

The authors received no funding from an external source.